\documentclass[10pt]{article}
\usepackage{amssymb}         
\usepackage{amsmath}    
\usepackage{graphicx}

\newcommand{\Xb}{{\mathbf{X}}}
\newcommand{\ab}{{\mathbf{a}}}
\newcommand{\qb}{{\mathbf{q}}}
\newcommand{\rb}{{\mathbf{r}}}
\newcommand{\xib}{{\mathbf{\xi}}}
\newcommand{\Eb}{{\mathbf{E}}}
\newcommand{\Rb}{{\mathbf{R}}}

\newcommand{\Xcal}{\mathcal{X}}
\newcommand{\Hcal}{\mathcal{H}}
\newcommand{\Acal}{\mathcal{A}}
\newcommand{\Gcal}{\mathcal{G}}
\newcommand{\Fcal}{\mathcal{F}}

\newcommand{\Pcal}{\mathcal{P}}
\newcommand{\Lcal}{\mathcal{L}}

\newcommand{\Qcal}{\mathcal{Q}}
\newcommand{\Dcal}{\mathcal{D}}

\DeclareMathOperator*{\argmin}{argmin}

\newtheorem{thm}{Theorem}
 
\newtheorem{lemma}[thm]{Lemma} 
 
\newtheorem{proposition}[thm]{Proposition} 
\newtheorem{remark}[thm]{Remark}
\newtheorem{corollary}[thm]{Corollary}

\begin{document}

\title{Strong Consistency of Prototype Based Clustering in Probabilistic Space}
\author{\textsc{Vladimir Nikulin\thanks{Email: vnikulin.uq@gmail.com}}    \\ 
Department of Mathematics, University of Queensland \\
Brisbane, Australia
\and
\textsc{Geoffrey J. McLachlan\thanks{Email: gjm@maths.uq.edu.au}}  \\ 
Department of Mathematics, University of Queensland \\
Brisbane, Australia
}
\date{}
\maketitle
\thispagestyle{empty}

\begin{abstract}
   In this paper we formulate in general terms an approach to prove strong consistency of
   the \textit{Empirical Risk Minimisation} 
   inductive principle applied to the prototype or distance based clustering.
   This approach was motivated by the Divisive Information-Theoretic 
   Feature Clustering model in 
   probabilistic space with Kullback-Leibler divergence which may be regarded
   as a special case within the \textit{Clustering Minimisation} framework. 
   Also, we propose clustering regularization restricting creation of additional 
   clusters which are not significant or are not essentially different comparing with
   existing clusters.
\end{abstract}
\section{Introduction}
 Clustering algorithms group data according to the given criteria. 
 For example, it may be a model based on \textit{Spectral Clustering} 
 \cite{NgJorWei02} or \textit{Prototype Based} model \cite{HinKei03}.

In this paper we consider a \textit{Prototype Based} approach which may be 
 described as follows.
 Initially, we have to choose $k$ \textit{prototypes}.
 Corresponding empirical clusters will be defined in accordance to the criteria of the 
 nearest prototype measured by the distance $\Phi$. 
 Respectively, we will generate initial $k$ clusters.
 As a second Minimisation step we will recompute 
 \textit{cluster centers} or \textit{$\Phi$-means} \cite{CuGoMar97} 
 using data strictly from the corresponding clusters. Then, we can repeat 
 Clustering step using new prototypes 
 obtained from the previous step as a cluster centers.
 Above algorithm has descending property. Respectively, it will reach
 local minimum in a finite number of steps.

 Pollard \cite{Pol81} demonstrated that the classical $K$-means algorithm in 
 $\mathbb{R}^m$ with squared loss function satisfies 
 the Key Theorem of Learning Theory \cite{Vap95}, p.36, 
 \textit{``the minimal empirical risk must converge to the minimal actual risk''}.

 A new  clustering algorithm in probabilistic space $\Pcal^m$ was 
 proposed in \cite{DhiMalKum03}. It provides an attractive approach 
 based on the Kullback Leibler divergence. The above methodology 
 requires a general 
 formulation and framework which we will 
 present in the following Section~\ref{sec:ClustProbSpace}. 

 Section~\ref{sec:defin} extends the methodology of \cite{Pol81} in 
 order to cover the case of $\Pcal^m$ with Kullback Leibler divergence.
 Using the results and definitions of the Section~\ref{sec:defin}, 
 we investigate relevant properties of $\Pcal^m$ in the 
 Section~\ref{sec:probframe} and prove   
 a strong consistence of the Empirical Risk Minimisation inductive principle.

Determination of the number of clusters $k$ represents an important 
 problem. For example, \cite{HamElk03} proposed the $G$-means algorithm which is based on the
 $Gaussian$ fit of the data within particular cluster. Usually 
 attempts to estimate the number of Gaussian clusters will lead 
 to a very high value of $k$ \cite{ZhGh03}.
 Most simple criteria such as $AIC$ (\textit{Akaike Information Criterion} \cite{Akaike78}) 
 and $BIC$ (\textit{Bayesian Information Criterion} \cite{Sch78}, \cite{FrRaf98}) 
 either overestimate or underestimate the number of clusters, which severely limits 
 their practical usability. 
 We introduce in Section~\ref{ssec:detnumclust} special clustering regularization. 
 This regularization will restrict creation of a new cluster which is not big enough and  
 which is not sufficiently different comparing with existing clusters.

\section{Prototype Based Approach}  \label{sec:ClustProbSpace}

In this paper we will consider a sample of i.i.d. observations 
 $\textbf{X} := \{x_1, \ldots, x_n\}$ 
 drawn from probability space $(\Xcal, \Acal, \mathbb{P})$
 where probability measure $\mathbb{P}$ is assumed to be unknown.

 Key in this scenario is an encoding problem.
 Assuming that we have a codebook $\Qcal \in \Xcal^k$ with 
 \textit{prototypes} $q(c)$ indexed by the code $c = 1, \ldots ,k$,
 the aim is to encode any $x \in \mathcal{X}$ by some $q(c(x))$ such that the
 distortion between $x$ and $q(c(x))$ is minimized:
\begin{equation} \label{eq:code}
   c(x) := argmin_c \Lcal (x, q(c))
\end{equation}
 where $\Lcal(\cdot,\cdot)$ is a loss function. 

 Using criterion (\ref{eq:code}) we split empirical data into $k$ clusters.
 As a next step we compute the \textit{cluster center} specifically for any 
 particular cluster in order to minimise overall distortion error.

 We estimate actual distortion error
\begin{align}
  \label{eq: cda}
  \Re^{(k)}[\Qcal] := \Eb \hspace{0.1in} \Lcal (x, \Qcal)
\end{align}
by the empirical error
\begin{align}
  \label{eq: cda-emp}
  \Re^{(k)}_\mathrm{emp}[\Qcal] := 
  \frac{1}{n} \sum_{t=1}^n \Lcal (x_t, \Qcal)
\end{align}
where $\Lcal(x, \Qcal) := \Lcal(x, q(c(x)))$.

The following Theorem, which may be proved similarly to the 
Theorems 4 and 5 of \cite{DhiMalKum03}, 
formulates the most important descending and convergence properties within the 
\textit{Clustering Minimisation} (CM) framework:
\begin{thm} \label{th: rpm}
  The $CM$-algorithm includes 2 steps:
  \textbf{Clustering Step}: recompute $c(x)$ according to (\ref{eq:code}) 
  for a fixed prototypes from the given codebook $\Qcal$, 
  which will be updated as a cluster centers from the next step, 

  \textbf{Minimisation Step}: recompute cluster centers for a fixed mapping $c(x)$ or
   minimize the objective function (\ref{eq: cda-emp}) over $\Qcal$, and

  1) monotonically decreases the value of the objective function (\ref{eq: cda-emp});

  2) converges to a local minimum in a finite number of steps 
  if Minimisation Step has exact solution.
\end{thm}

We define an optimal actual codebook  $\overline{\Qcal}$ by the following condition:
\begin{equation}  \label{eq: expect}
   \Re^{(k)}(\overline{\Qcal}) := \underset{\Qcal \in \Xcal^k}{inf} 
   \Re^{(k)}(\Qcal).
\end{equation} 

The following relations are valid
\begin{equation}
\label{eq: conv1} 
     \Re^{(k)}_\mathrm{emp}[\Qcal_n] 
     \le \Re^{(k)}_\mathrm{emp}[\overline{\Qcal}];  \hspace{0.1in}  
     \Re^{(k)}_\mathrm{emp}[\overline{\Qcal}] \Rightarrow 
     \Re^{(k)}[\overline{\Qcal}] \hspace{0.1in} a.s.    
\end{equation}
where $\Qcal_n$ is an optimal empirical codebook:
\begin{equation}  \label{eq: defn2}
   \Re_{\mathrm{emp}}^{(k)}(\Qcal_n) := 
   \underset{\Qcal \in \Xcal^k}{inf} { \{ \Re_{\mathrm{emp}}^{(k)}(\Qcal)} \}.
\end{equation}
The main target is to demonstrate asymptotical (\textit{almost sure}) convergence
\begin{equation}  \label{eq: defn3}
   \Re_{\mathrm{emp}}^{(k)}(\Qcal_n) \Rightarrow 
   \Re^{(k)}[\overline{\Qcal}] \hspace{0.1in} a.s. 
   \hspace{0.1in} \left( n \rightarrow \infty \right).
\end{equation}
In order to prove (\ref{eq: defn3}) we define in
Section~\ref{sec:defin} general model which has direct relation to 
the model in probabilistic space $\Pcal^m$ with
with $KL$ divergence \cite{DhiMalKum03}.

The proof of the main result which is formulated in the 
Theorem~\ref{th: uconv} includes two steps:
\renewcommand{\theenumi}{\arabic{enumi}}
\renewcommand{\labelenumi}{(\theenumi)}
\begin{enumerate}
\item by Lemma~\ref{th: mlemma} we prove existence of $n_0$ such 
  that $\Qcal_n \subset \Gamma$ for all $n \geq n_0$ 
  where subset $\Gamma \subset \Xcal$ satisfies condition:
   $\Lcal(x,q) < \infty$ for all $x \in \Xcal, q \in \Gamma$; and 

\item by Lemma~\ref{th: slemma} we prove 
  (under some additional constraints of general nature)  
\begin{equation}  \label{eq: uslln}
  \underset{\Qcal \in \Gamma^k}{sup} |\Re^{(k)}_\mathrm{emp}[\Qcal] -
  \Re^{(k)}[\Qcal] | \Rightarrow 0 \hspace{0.1in} a.s.
\end{equation}
\end{enumerate}

\section{General Theory and Definitions}  \label{sec:defin}

In this section we employ some ideas and methods proposed in \cite{Pol81}, and which
cover the case of $\mathbb{R}^m$ with loss function
$\Lcal(x,q):=\varphi(\|x-q\|)$ where $\varphi$ is a strictly increasing function.

Let us assume that the following structural representation with 
$\mathbb{P}$-integrable vector-functions $\xi$ and $\eta$ is valid
\begin{equation}  \label{eq: loss}
  \Lcal(x,q) := \sum_{i=0}^m \xi_i(x) \cdot 
  \eta_i(q) = \langle \xi(x), \eta(q) \rangle \ge 0
  \hspace{0.1in} \forall x,q \in \Xcal.
\end{equation}
Let us define subsets of $\Xcal$ as extensions of the empirical clusters:
\begin{center}
$\Xcal_c(\Qcal) := \left\{ x \in \Xcal: \hspace{0.1in} 
c = \argmin_i \mathcal{L} (x,q(i)) \right\},$
\end{center}
\begin{center}
$\Xcal= \cup_{c=1}^k \Xcal_c(\Qcal), 
\Xcal_i(\Qcal) \cap \Xcal_c(\Qcal) = \emptyset, i \ne c$.
\end{center}
Then, we can re-write (\ref{eq: cda}) as follows
\begin{equation}  \label{eq: tlln}
  \Re^{(k)}[\Qcal] :=  \sum_c \langle \xib(\Xcal_c), \eta(q(c)) \rangle 
\end{equation}
where $\xib(A) := \int_A \xi(x) \mathbb{P}(dx), A \in \Acal$.

We define a ball with radius $r$ and a corresponding reminder in $\Xcal$ 
\begin{subequations}
\begin{align}
\label{eq:ball} 
     B(r) = \{ q \in \Xcal : \Lcal (x, q) \le r, 
     \hspace{0.1in} \forall x \in \Xcal \},      \\
\label{eq: remn}
     T(r) = \Xcal \setminus B(r), \hspace{0.1in} r \geq \rb_0,  \\
\label{eq: rem0}
\rb_0 = \inf \{ r \geq 0: B(r) \neq \emptyset \}.       
\end{align} 
\end{subequations}  

The following properties are valid
\begin{equation}  \label{eq: prpt}
  \langle \xib(A_1) - \xib(A_2), \eta(q) \rangle \geq 0
\end{equation}
for all $q \in \Xcal$ and any $A_1, A_2 \in \Acal: A_2 \subset A_1$;  
\begin{equation}  \label{eq: prpt1}
 \langle \xib(\Xcal), \eta(q) \rangle \leq r \hspace{0.1in} \forall q \in B(r).
\end{equation}

Suppose, that 
\begin{equation}    \label{eq: gcnd6}
   \mathbb{P}(T(U)) \underset{U \rightarrow \infty}{\longrightarrow} 0.
\end{equation}

The following distances will be used below:
\begin{equation}   \label{eq: metr1} 
    \rho(A_1, A_2) := \underset{a_1 \in A_1}{inf} 
    \underset{a_2 \in A_2}{inf} \Lcal (a_1, a_2), A_1, A_2 \in \Acal;   
\end{equation}
\begin{equation}    \label{eq: metr2}
    \mu(A_1,A_2) = \underset{a_1 \in A_1}{inf} 
    \underset{a_2 \in A_2}{sup} \Lcal(a_1, a_2), A_1, A_2 \in \Acal. 
\end{equation}

Suppose, that 
\begin{equation}   \label{eq: gcnd4}
   \rho(B(r),T(U)) \underset{U \rightarrow \infty}{\longrightarrow} \infty
\end{equation}
for any fixed $\rb_0 \leq r < \infty$.

\textbf{Remark 1} \hspace{0.02in} We assume that 
\begin{equation}  \label{eq: gcnd5}
   T(U) \neq \emptyset
\end{equation}
for any fixed $U: \rb_0 \leq U < \infty$, alternatively, the following below 
Lemma~\ref{th: mlemma} become trivial.

\setcounter{thm}{0}
\begin{lemma} \label{th: mlemma} Suppose, that the structure of the loss function $\Lcal$ 
 is defined in (\ref{eq: loss}) under condition (\ref{eq: gcnd4}).
 Probability distribution $\mathbb{P}$ satisfies condition (\ref{eq: gcnd6}) 
 and the number of
 clusters $k \geq 1$ is fixed. 
 Then, we can select large enough radius \hspace{0.05in} 
 $Z: 0 < Z < \infty$ and $n_0 \ge 1$ such that all components of the optimal empirical
 codebook $\Qcal_n$ defined in (\ref{eq: defn2}) will be within the ball $B(Z)$:
 $\Qcal_n \subset B(Z)$ if sample size is large enough: $\forall n \geq n_0$.
\end{lemma}
\textit{Proof:} Existence of the element $\ab \in \Xcal$ such that
\begin{equation}  \label{eq: defn}
   D_{\ab} = \Re^{(1)}(\{\ab\}) = 
   \langle \xi(\Xcal), \eta(\ab) \rangle < \infty
\end{equation}
follows from (\ref{eq: prpt1}) and (\ref{eq: gcnd6}).  

Suppose that
\begin{equation}  \label{eq: cond1}
    \mathbb{P}(B(r)) = P_0 >0,  \hspace{0.07in} r \geq \rb_0.
\end{equation}
We can construct $B(V)$ in accordance with condition (\ref{eq: gcnd4}) and 
(\ref{eq: gcnd5}):
\begin{equation}  \label{eq: cond1}
  V = \inf{ \{ v > r: \rho(B(r), T(v))   
  \geq \frac{D_\ab + \epsilon}{P_0} \}}, \hspace{0.1in} \epsilon > 0.
\end{equation}
Suppose, there are no empirical prototypes within $B(V)$. 
Then, in accordance with definition (\ref{eq: cond1}) 
\begin{equation*}  
   \Re^{(k)}_\mathrm{emp}[\Qcal_n] \geq D_{\ab} + \epsilon > D_{\ab} 
   \hspace{0.07in} \forall n >0.
\end{equation*}
Above \textit{contradicts} to (\ref{eq: defn}) and (\ref{eq: conv1}). 
Therefore, at least one prototype
from $\Qcal_n$ must be within $B(V)$ if $n$ is large enough (this fact 
is valid for $\overline{\Qcal}$ as well). 
Without loss of generality we assume that
\begin{equation}  \label{eq: inside}  
   q(1) \in B(V).
\end{equation} 
The proof of the Lemma has been completed in the case if $k = 1$.
Following the method of mathematical induction, suppose, that $k \ge 2$ and
\begin{equation}  \label{eq: optim}
   \Re^{(k-1)}(\overline{\Qcal}) - 
   \Re^{(k)}(\overline{\Qcal}) \ge \varepsilon > 0. 
\end{equation} 
Then, we define a ball $B(U)$ by the following conditions
\begin{equation}  \label{eq: cond2}
  U = \inf{ \{ u > V: \underset{q \in B(V)}{sup} 
  \langle \xi(T(u)), \eta(q) \rangle < \varepsilon \}}.
\end{equation}
Existence of the $U: V < U < \infty$ in (\ref{eq: cond2}) follows from 
(\ref{eq: prpt1}) and (\ref{eq: gcnd6}).

By definition of the distance $\mu$ and ball $B(V)$ 
\begin{equation}  \label{eq: cond3}
   0 < \Dcal(U,V) = \mu(T(U), B(V)) \leq V < \infty.
\end{equation}
Now, we can define reminder $T(Z) \neq \emptyset$ in accordance with condition 
(\ref{eq: gcnd4}):
\begin{equation}  \label{eq: cond4}
   Z = \inf{ \{ z > U: \rho(B(U), T(z)) \ge \Dcal(U,V) \}}.
\end{equation}
Suppose, that there is at least one prototype within $T(Z)$,
for example, $q(2) \in T(Z)$.
On the other hand, we know about (\ref{eq: inside}). 
Let us consider what will happen if we will remove $q(2)$ from the optimal empirical
codebook $\Qcal_n$ (the case of optimal actual risk 
$\overline{\Qcal}$ may be considered similarly) and will replace it by $q(1)$:
\renewcommand{\theenumi}{\arabic{enumi}}
\renewcommand{\labelenumi}{(\theenumi)}
\begin{enumerate}
\item as a consequence of (\ref{eq: cond3}) and (\ref{eq: cond4}) all empirical data within 
     $B(U)$ are closer to $q(1)$ anyway, 
     means the data from $B(U)$ will not increase empirical (or actual) 
risk (\ref{eq: cda-emp});
\item by definition, $\Xcal = B(U) \cup T(U), B(U) \cap T(U) = \emptyset$ 
     and in accordance with the condition (\ref{eq: cond2}) an empirical risk
     increases because of the data within $T(U)$ must be strictly less
     compared with $\varepsilon$ for all large enough $n \geq n_0$   
     (actual risk increase will be strictly less compared with $\varepsilon$ 
     for all $n \geq 1$). 
\end{enumerate}
Above \textit{contradicts} to the condition (\ref{eq: optim}) and (\ref{eq: conv1}). 
Therefore,
all prototypes from $\overline{\Qcal}$ must be within 
$\Gamma = B(Z)$ for all $n \geq 1$, and 
$\Qcal_n \subset \Gamma$ if $n$ is large enough. $\blacksquare$ 

\subsection{Uniform Strong Law of Large Numbers (SLLN)}

Let $\Fcal$ denote the family of $\mathbb{P}$-integrable functions on $\Xcal$.

A sufficient condition for \textit{uniform SLLN} (\ref{eq: uslln}) 
is: for each $\delta > 0$ there exists
a \textit{finite} class $\Fcal_{\delta} \in \Fcal$ such that to 
each $\Lcal \in \Fcal$ there are
functions $\underline{\Lcal}$ and $\overline{\Lcal} \in \Fcal_{\delta}$ with the following 
2 properties:
\begin{center}
$\underline{\Lcal}(x) \le \Lcal(x) \le \overline{\Lcal}(x)$ for all $x \in \Xcal$; 
 \hspace{0.1in}
$\int_{\Xcal} \left( \overline{\Lcal}(x) - 
 \underline{\Lcal}(x) \right) \mathbb{P}(dx) \le \delta$.
\end{center}
We shall assume here existence of the function $\varphi$ such that
\begin{equation}  \label{eq: cmpct}
      \| \eta(q) \| \leq \varphi(Z) < \infty
\end{equation}
for all $q \in B(Z)$ where $\rb_0 \leq Z < \infty$.

\begin{lemma} \label{th: slemma}
  Suppose that the number of clusters $k$ is fixed and the loss function 
  $\Lcal$ is defined by (\ref{eq: loss}) under conditions 
  (\ref{eq: cmpct}) and
\begin{equation}  \label{eq: frst}
 \| \xi(x) \| \le \Rb < \infty 
 \hspace{0.1in} \forall x \in \Xcal.
\end{equation}
 Then, the asymptotical relation (\ref{eq: uslln}) is valid for any $\Gamma = B(Z),
 \rb_0 \leq Z \leq \infty$.
\end{lemma}
\textit{Proof:} Let us consider the definition of 
Hausdorff metric $\Hcal$ in  $\mathbb{R}^{m+1}$: 
 $$\Hcal(A_1,A_2) = \underset{a_1 \in A_1}{sup} 
    \underset{a_2 \in A_2}{inf} \| a_1 - a_2 \|,$$
and denote by $\Gcal$ a subset in $\mathbb{R}^{m+1}$ which was obtained 
from $\Gamma$ as a result of $\eta$-transformation.
According to the condition (\ref{eq: cmpct}), 
$\Gcal$ represents a compact set. It means, existence of a finite
subset $\Gcal_{\delta}$ for any $\delta >0$ 
such that $\Hcal(\Gcal,\Gcal_{\delta}) \leq \frac{\delta}{2 \Rb}$ 
where $\Rb$ is defined in (\ref{eq: frst}).
We denote by $\Gamma_{\delta} \subset \Gamma$ subset which corresponds 
to $\Gcal_{\delta} \subset \Gcal$ according to the $\eta$-transformation.
Respectively, we can define transformation 
(according to the principle of the nearest point)
$f_{\delta}$ from $\Gamma$ to $\Gamma_{\delta}$,
and $\Qcal_{\delta} = f_{\delta}(\Qcal)$ where 
closeness may be tested independently for any particular component of $\Qcal$, 
that means absolute closeness.

In accordance with the Cauchy-Schwartz inequality, the following relations take place
\begin{equation*}
\underline{\Lcal} = \Lcal(x,\Qcal_{\delta})-\frac{\delta}{2} \leq  \Lcal(x,\Qcal) \leq  
 \Lcal(x,\Qcal_{\delta})+\frac{\delta}{2} = \overline{\Lcal} 
 \hspace{0.05in} \forall x \in \Xcal.
\end{equation*}
Finally, $\int_{\Xcal} \left( \overline{\Lcal}(x,\Qcal_{\delta}) -
  \underline{\Lcal}(x,\Qcal_{\delta}) \right) 
  \mathbb{P}(dx) \le \delta$  where $\Qcal_{\delta} \in \Gamma^k_{\delta}$ is the 
absolutely closest codebook for the arbitrary $\Qcal \in \Gamma^k$. $\blacksquare$

\section{A Probabilistic Framework}  \label{sec:probframe}

Following \cite{DhiMalKum03}, we assume that the probabilities 
$p_{\ell t}=P(\ell | x_t),
\sum_{\ell =1}^m p_{\ell t} =1, t=1, \ldots ,n$,
represent relations between observations $x_t$ and attributes or classes 
$\ell = 1, \ldots ,m, m \geq 2$. 

Accordingly, we will define probabilistic space $\Pcal^m$ of all $m$-dimensional probability
vectors with \textit{Kullback-Leibler} ($KL$) divergence:
\begin{equation*}
KL(v, u) := \sum_{\ell} v_{\ell} \cdot 
\log{ \frac{v_{\ell}}{u_{\ell}}} = 
\langle v, \log{ \frac{v}{u}} \rangle \hspace{0.05in} v, u \in \Pcal^m.
\end{equation*}
\textbf{Graphical Example}. Figure 1(a) illustrates 
first two coordinates of the synthetic data in $\Pcal^3$. 
Third coordinate is not necessary because it is a function of the first two coordinates.
\begin{figure}
\includegraphics[scale=.41]{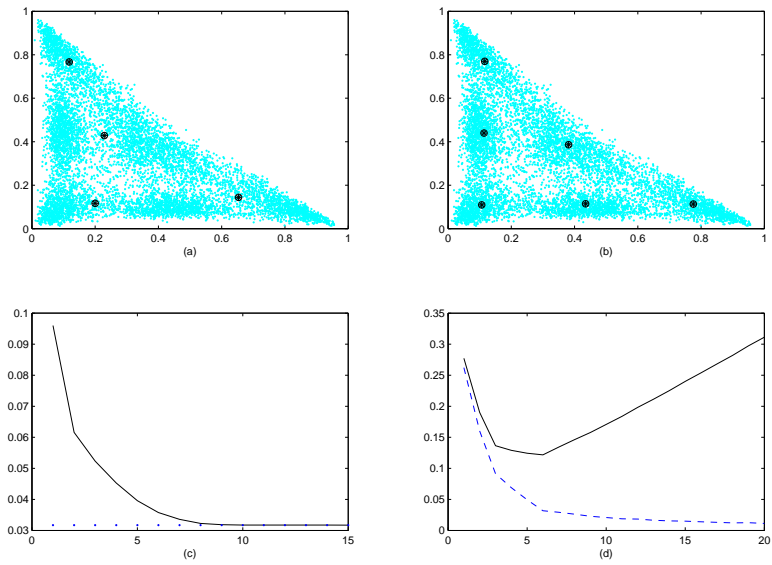}
  \caption{(a-b) 3D Probability data with 4 and 6 cluster centers, $n=8000$;
  (c) convergence of the $CM$ algorithm based on 
  $KL$ divergence in the case of 6 clusters; 
  (d) behavior of the empirical error (\ref{eq: cda-emp}) (blue, dashed) 
  and empirical error
  with cost term (\ref{eq:rpm1}) where $\alpha = 0.1, \beta = 0.03$.}
\label{fig: figure1}
\end{figure}

\textbf{Remark 2} \hspace{0.02in} As it was demonstrated in 
\cite{DhiMalKum03}, cluster centers $q_c$ in the space $\Pcal^m$
with $KL$-divergence must be computed using $K$-means:
\begin{align}
  \label{eq:kmns}
  q_c = \frac{1}{n_c} \sum_{x_t \in \Xb_c} p_t
\end{align} 
where $c(x_t)=c$ if $x_t \in \Xb_c$ and
$n_c = \#\Xb_c$ is the number of observations 
in the cluster $\Xb_c, c=1, \ldots ,k$, 
$p_t = \{ p_{1t}, \ldots , p_{mt} \}, q_c = \{ q_{1t}, \ldots , q_{mt} \}$.

In difference to the model of \cite{Pol81} in $\mathbb{R}^{m}$, 
the structure (\ref{eq: loss}) 
covers an important case of $\Pcal^m$ with $KL$-divergence:
\begin{align}     \label{eq:entrp}
  \xi_0(v) = \sum_{\ell=1}^m v_{\ell} \log{v_{\ell}}; \hspace{0.1in} 
  \xi_{\ell}(v) = v_{\ell};
\end{align}
\begin{center}
  $\eta_0(u) = 1; \hspace{0.1in} 
  \eta_{\ell}(u) = - \log{u_{\ell}}, \ell = 1, \ldots ,m.$
\end{center}
\textbf{Definition}. We will call element $v \in \Pcal^m$  
as 1) \textit{uniform center} if 
$v_{\ell} = \frac{1}{m}, \ell =1, \ldots ,m$; 
as 2) \textit{absolute margin} if $\min_{\ell}{v_{\ell}} =0$.
\setcounter{thm}{0}
\begin{proposition} \label{th: lem5}  The ball $B(Z) \subset \Pcal^m$ 
contains only one element named as uniform center in the case if 
$Z = \rb_0 = \log{(m)}$, 
and $B(Z) = \emptyset$ if $Z < \rb_{0}$.
\end{proposition}
\textit{Proof:} Suppose, that $u$ is a uniform center. Then,
$KL(v, u) = \sum_{i=1}^m v_i \log{v_i} + \log{m} \leq \log{m}$ 
for all $v \in \Pcal^m$.
In any other case, one of the components of $u$ must be less than 
$\frac{1}{m}$. Respectively, we can select
corresponding component of the probability vector $v$ as $1$.
Therefore, $KL(v, u) > \log{(m)}$ and $\rb_0 = \log{(m)}$. $\blacksquare$

\setcounter{thm}{2}
\begin{lemma} \label{th: lemma3} The 
$KL$ divergence in probabilistic space $\Pcal^m$ always satisfies condition
  (\ref{eq: frst}) where vector-function $\xi$ is expressed by (\ref{eq:entrp})
  with the following upper bounds:
\begin{equation*}
    | \xi_0(v) | \leq \log{(m)}; \hspace{0.1in} 
    | \xi_{\ell} (v) | \leq 1, \ell = 1, \ldots ,m, \hspace{0.1in} 
    \hspace{0.1in} \forall v \in \Pcal^m.
\end{equation*}
\end{lemma}
\begin{lemma} \label{th: lemma4} The following relations are valid 
 in $\Pcal^m$ 
\renewcommand{\theenumi}{\arabic{enumi}}
\renewcommand{\labelenumi}{(\theenumi)}
\begin{enumerate}
  \item $min_{\ell} \{ u_{\ell} \} < e^{ - r}$ for all $u \in T(r)
   \hspace{0.1in} \forall r \geq \rb_0$;
  \item $u_{\ell} \geq e^{-r}$ for all $\ell=1, \ldots ,m$, and any $u \in B(r)
   \hspace{0.1in} \forall r \geq \rb_0$.
\end{enumerate}  
\end{lemma}
\textit{Proof:} As far as $\Pcal^m = B(r) \cup T(r), B(r) \cap T(r) = \emptyset$,
the first statement may be regarded as consequence of the second. Suppose,
that $u \in B(r)$ and $u_1 = e^{-r-\varepsilon}, \varepsilon > 0$. Then, 
we can select $v_1 =1$, and $KL(v, u) = r + \varepsilon > r$ - 
\textit{contradiction}. $\blacksquare$

\setcounter{thm}{0}
\begin{corollary}    \label{axm:kldivps} The $KL$ divergence in  
  $\Pcal^m$ always satisfies conditions (\ref{eq: gcnd4}) and
\begin{equation*}
  -\log{(m)} + Z \cdot e^{- r} < \rho(B(r), T(Z)) \leq 
  e^{-r} \cdot \left( Z - r \right) + \left( 1 - e^{-r} \right) 
  \log{ \frac{1-e^{-r}}{1-e^{-Z}}} 
\end{equation*}
for all \hspace{0.1in} $\rb_0 \leq r < Z$ where the distance $\rho$ 
is defined in (\ref{eq: metr1}).
\end{corollary}
\textit{Proof:} Suppose, that $v \in B(r)$ and $u \in T(Z)$. Then,
$-\sum_{i=1}^m v_i \log{(u_i)} > Z \cdot e^{-r}$ for all 
\hspace{0.1in} $r: \rb_0 \leq r < Z$. On the other hand, the entropy
$H(v) = - \sum_{i=1}^m v_i \log{(v_i)}$ may not be smaller comparing with 
$\log{(m)}$. The low bound is \textit{proved}.
In order to prove the upper bound we shall suppose without loss of 
generality that $v_1 = e^{-r}, u_1 = e^{-Z}$, and all other components are
proportional. $\blacksquare$

\setcounter{thm}{1}
\begin{thm} \label{th: uconv} Suppose that probability measure
   $\mathbb{P}$ satisfies condition (\ref{eq: gcnd6}) in probabilistic space $\Pcal^m$ 
   with $KL$ divergence and number of clusters $k$ is fixed.
   Then, the minimal empirical error (\ref{eq: defn2}) will
   converge to the minimal actual error (\ref{eq: expect}) with probability 1 or a.s. 
\end{thm}
\textit{Proof:} Follows directly from the 
Lemmas~\ref{th: mlemma}, \ref{th: slemma}, \ref{th: lemma3} and \ref{th: lemma4}. 

\textbf{Remark 3} \hspace{0.02in} Condition (\ref{eq: gcnd6}) will not be valid if and only if 
a probability of the subset of all absolute margins is strictly positive. 
Note that in order to avoid any problems with consistency 
we can generalise definition 
of $KL$-divergence using special smoothing parameter $0 \leq \theta \leq 1$:
$$KL_{\theta}(v,u) = KL(v_{\theta}, u_{\theta})$$
where $v_{\theta} = \theta v + (1-\theta)v_0$ and 
$u_{\theta} = \theta u + (1-\theta)v_0$, $v_0$ is uniform center.

\section{Clustering Regularization} \label{ssec:detnumclust}

Let us introduce the following definitions:
\begin{equation*}
 \qb_c := \frac{1}{n_c} \sum_{x_t \in \Xb_c} p_t; \hspace{0.1in} 
 \qb := \frac{1}{n} \sum_{x_t \in \Xb} p_t = \sum_{c=1}^k p_c \cdot \qb_c; 
\end{equation*} 
\begin{equation*}
 H(\qb_c) :=  - \langle \qb_c, \log{\qb_c} \rangle; \hspace{0.1in} 
 H(\Xb) := \frac{1}{n} \sum_{x_t \in \Xb} H(x_t) 
\end{equation*}
where $p_c = P(\Xb_c) = \frac{n_c}{n}, c =1, \ldots ,k,$ and 
$H(x_t) = - \langle p_t, \log{p_t} \rangle$.

We define in this section a regularisation to restrict usage of 
unnecessary clusters. This regularisation is based on the following two conditions: 

\renewcommand{\theenumi}{C\arabic{enumi}}
\renewcommand{\labelenumi}{\theenumi)}
\begin{enumerate}
  \item $p_c \geq \alpha>0, c=1, \ldots ,k$ (\textit{significance of any particular cluster}); 
  \item $KLS(\qb_i, \qb_c) := KL(\qb_i, \qb_c) + KL(\qb_c, \qb_i) \geq \beta > 0$ 
  (\textit{difference between any 2 clusters $i$ and $c, i \ne c$}). 
\end{enumerate}

According to \cite{HinKei03}, if more prototypes are used for the 
$k$-means clustering, the algorithm splits clusters, which means that 
it represents a single cluster by more than one prototype. 
The following Proposition~\ref{axm:ncx2} considers clustering 
procedure in an inverse direction.
\setcounter{thm}{1}
\begin{proposition}    \label{axm:ncx2}
The following representations are valid
$$\Re_\mathrm{emp}^{(k)} = -H(\Xb) + \sum_c p_c H(\qb_c); \hspace{0.05in} 
\Re_\mathrm{emp}^{(1)} - \Re_\mathrm{emp}^{(k)} = 
\sum_{c=1}^k p_c KL(\qb_c, \qb), \hspace{0.05in} \forall n \geq k \geq 1.$$
\end{proposition}
\textit{Proof:} In accordance with above definitions
$$ \Re_\mathrm{emp}^{(k)} = - H(\Xb) - 
 \sum_c \frac{n_c}{n} \sum_{x_t \in \Xb_c} \frac{1}{n_c} 
 \langle p(x_t), \log{\qb_c} \rangle, \hspace{0.05in} \forall k \geq 1,$$
and
$$\Re_\mathrm{emp}^{(1)} - \Re_\mathrm{emp}^{(k)} =
 \sum_c p_c H(\qb_c) -  H(\qb) =     \\
\sum_c p_c \langle \qb_c, \log{\qb_c} - \log{\qb} \rangle$$
where the second equation follows directly from the first one. $\blacksquare$
\setcounter{thm}{0}
\begin{corollary}    \label{axm:ncx4}
Assuming that we merge first $\tau$ clusters, $1 \leq \tau \leq k$, 
the following relation is valid 
\begin{equation}   \label{eq: dncln1}
 \Re_\mathrm{emp}^{(k-\tau+1)} - \Re_\mathrm{emp}^{(k)} \leq 
 \sum_{c=1}^{\tau} p_c KL(\qb_c, \widehat{\qb}_{\tau}), \hspace{0.1in} 
  \widehat{\qb}_{\tau} = \frac{\sum_{c=1}^{\tau} p_c \qb_c}{\sum_{c=1}^{\tau} p_c}.
\end{equation}
\end{corollary}
\textbf{Remark 4} \hspace{0.02in} 
First $\tau$ clusters were chosen in order to simplify notifications and without loss
of generality.

As a result of standard application of Jensen's inequality to (\ref{eq: dncln1})
we can formulate similar results in terms of particular differences between clusters.
\begin{corollary}    \label{axm:ncx5}
The following relation is valid
\begin{equation}  \label{eq:dncln2}
 \Re_\mathrm{emp}^{(k-\tau+1)} - \Re_\mathrm{emp}^{(k)} 
 \leq \frac{\sum_{i=1}^{\tau} \sum_{c=1}^{\tau} p_i p_c 
 KL(\qb_i, \qb_c)}{\sum_{c=1}^{\tau} p_c}
\end{equation}
for any $n \geq k \geq 2$.
\end{corollary}
As a direct consequence of (\ref{eq:dncln2}), we derive formula 
for the case of two clusters indexed by $i$ and $c$:
\begin{equation}
  \label{eq:mgij}
  \Re_\mathrm{emp}^{(k-1)} - \Re_\mathrm{emp}^{(k)}
  \leq \frac{p_i \cdot p_c \cdot
   KLS(q_i, q_c) }{p_i + p_c}.
\end{equation}

The coefficient $\frac{p_i \cdot p_c}{p_i + p_c}$ in 
(\ref{eq:mgij}) represents an increasing
function of probabilities $p_i$ and $p_c \geq \alpha$. 
Respectively, we form regularized empirical risk by including additional 
$cost$ term in (\ref{eq: defn2}): 
\begin{equation}  \label{eq:rpm1a}
  \Re^{(k)}_\mathrm{emp}[\Qcal_n] + C(k) 
\end{equation}
where
\begin{equation}  \label{eq:rpm1}
  C(k) = \frac{\alpha \cdot \beta \cdot k}{2}. 
\end{equation}
Minimizing above regularized empirical risk as a function of number of clusters 
$k$ we will make required selection of the clustering size 
(see Figures~\ref{fig: figure1}(d)).
 \setcounter{thm}{4}
 \begin{remark} Note a structural similarity between (\ref{eq:rpm1a}) 
 and  Akaike Information Criterion \cite{Akaike73} and \cite{Akaike78}, 
 which has different grounds. In accordance with AIC, the 
 empirical log-likelihood is greater compared with 
 the actual log-likelihood because 
 we use the same data in order to estimate the required parameters.
 Asymptotically, the bias represents a linear function of the number of 
 the used parameters.
 \end{remark}
 
\section{Concluding Remarks}

Cluster analysis, an unsupervised learning method \cite{WWong05}, is widely used 
to study the structure of the data when no specific response variable is specified. 
Recently, several new clustering algorithms (e.g., graph-theoretical clustering, 
model-based clustering) 
have been developed with the intention to combine and improve the features of traditional 
clustering algorithms. However, clustering algorithms are based on different assumptions, 
and the performance of each 
clustering algorithm depends on properties of the input dataset. Therefore, the 
winning clustering algorithm 
does not exist for all datasets, and the optimization of existing clustering algorithms 
is still a vibrant research area \cite{Belacel06}. 

Probabilistic space with $KL$-divergence represents an essentially different case 
compared with Euclidean space with standard squared metric. 
In this paper we considered an illustration with a simple synthetic example. 
However, many real-life 
datasets may be transferred into probabilistic space as a result of the proper normalisation. 
For example, we know that all elements of the colon 
dataset\footnote{http://microarray.princeton.edu/oncology/affydata/index.html} are strictly 
positive. We can normalise any row of the colon matrix (which has interpretation as a gene) 
by division by the sum of the corresponding elements. As a next step, we can apply 
the model of Section~\ref{sec:probframe} in order to reduce dimensionality 
of the gene expression data. 
This analysis has an important role to play in the discovery, validation 
and understanding of various classes and subclasses of cancer \cite{Marron08}.

\end{document}